\documentclass[pdflatex,sn-mathphys-num]{sn-jnl}

\usepackage{makecell}
\usepackage{graphicx}%
\usepackage{multirow}%
\usepackage{amsmath,amssymb,amsfonts}%
\usepackage{amsthm}%
\usepackage{mathrsfs}%
\usepackage[title]{appendix}%
\usepackage{xcolor}%
\usepackage{textcomp}%
\usepackage{manyfoot}%
\usepackage{booktabs}%
\usepackage{algorithm}%
\usepackage{algorithmicx}%
\usepackage{algpseudocode}%
\usepackage{listings}%
\usepackage[table]{xcolor}
\usepackage{bm}

\theoremstyle{thmstyleone}%
\theoremstyle{thmstyletwo}%

\theoremstyle{thmstylethree}%

\begin{document}

\title[Article Title]{Investigating Long-term Training for Remote Sensing Object Detection}


\author[1]{\fnm{JongHyun} \sur{Park}}\email{citizen135@gm.gist.ac.kr}

\author*[1]{\fnm{Yechan} \sur{Kim}}\email{yechankim@gm.gist.ac.kr}

\author*[1]{\fnm{Moongu} \sur{Jeon}}\email{mgjeon@gist.ac.kr}

\affil*[1]{\orgdiv{Machine Learning and Vision Laboratory}, \orgname{Gwangju Institute of Science and Technology (GIST)}, \orgaddress{\city{Gwangju}, \postcode{61005}, \country{Republic of Korea}}}

\abstract{Recently, numerous methods have achieved impressive performance in remote sensing object detection, relying on convolution or transformer architectures.
    Such detectors typically have a feature backbone to extract useful features from raw input images.
    A common practice in current detectors is initializing the backbone with pre-trained weights available online.
    Fine-tuning the backbone is typically required to generate features suitable for remote-sensing images. 
    While the prolonged training could lead to over-fitting, hindering the extraction of basic visual features, it can enable models to gradually extract deeper insights and richer representations from remote sensing data.
    Striking a balance between these competing factors is critical for achieving optimal performance.
    In this study, we aim to investigate the performance and characteristics of remote sensing object detection models under very long training schedules, and propose a novel method named \underline{D}ynamic \underline{B}ackbone \underline{F}reezing (\textbf{DBF}) for feature backbone fine-tuning on remote sensing object detection under long-term training. 
    Our method addresses the dilemma of whether the backbone should extract low-level generic features or possess specific knowledge of the remote sensing domain, by introducing a module called `Freezing Scheduler' to manage the update of backbone features during long-term training dynamically.
    Extensive experiments on DOTA and DIOR-R show that our approach enables more accurate model learning while substantially reducing computational costs in long-term training.
    Besides, it can be seamlessly adopted without additional effort due to its straightforward design. The code is available at \url{https://github.com/unique-chan/dbf}.}




\maketitle

\section{Introduction}\label{sec1}

The rapid advancement of geospatial and deep learning technologies has revolutionized remote sensing object detection, enabling significant improvements in accuracy and efficiency. 
    Modern object detectors rely on a feature backbone such as convolutional neural networks and transformer architectures to extract meaningful representations from raw input images \cite{lee2022learning, lee2022perception, zhao2024orientedformer}. 
    These backbone models typically benefit from pre-trained weights on large-scale datasets such as ImageNet \cite{deng2009imagenet}, which serve as a foundation for learning low-level features like edges and textures. 
    Fine-tuning these pre-trained backbones with remote sensing datasets is a common practice to adapt to the unique characteristics of aerial and satellite imagery.

However, the process of fine-tuning under long-term training schedules introduces unique challenges and opportunities for remote sensing object detection. On the one hand, prolonged training can lead to overfitting, especially in feature backbones, hindering the extraction of fundamental visual features critical for generalization. On the other hand, such extended schedules provide a valuable opportunity to acquire domain-specific and enriched knowledge from data, potentially enhancing the model’s ability to detect intricate patterns unique to remote sensing imagery. For example, Beyer et al. \cite{beyer2022knowledge} empirically found that prolonged training with appropriate data augmentation strategies may be important to achieve better accuracy. This trade-off between preserving generic features and specializing for the domain presents a critical challenge in designing effective training strategies for feature backbones.

In addition, many modern object detection models are released with pre-trained weights based on natural image datasets such as ImageNet. To adapt these models to remote sensing tasks, fine-tuning is essential; however, as the model size increases, this process becomes increasingly resource-intensive. In particular, training large backbones from scratch or even fully fine-tuning them requires substantial GPU memory and computation time, which can hinder practical deployment in remote sensing scenarios \cite{kim2025nbbox, kim2025nsegment, kim2025unlocking}.

\begin{figure}[!t]
    \centering
    \includegraphics[width=10cm]{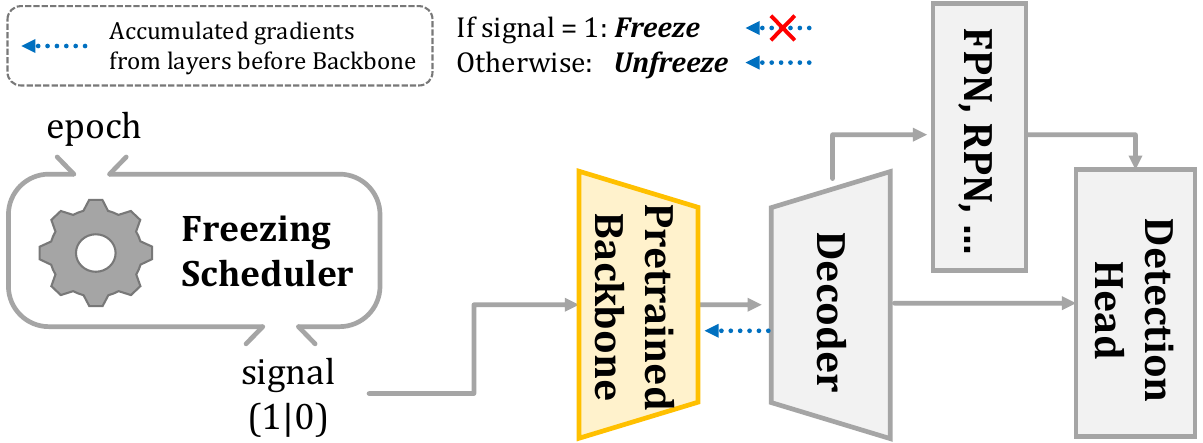}
    \caption{Illustration of training remote sensing object detectors with the proposed framework named Dynamic Backbone Freezing (\textbf{DBF}). \textbf{DBF} allows the backbone to receive knowledge from the downstream data when Freezing Scheduler sends a signal to open the backward route. This ensures robust prediction by preventing over-fitting while also significantly saving training costs.}
    \label{fig1}
\end{figure}

\begin{figure}[!t]
    \centering
    \includegraphics[width=12cm]{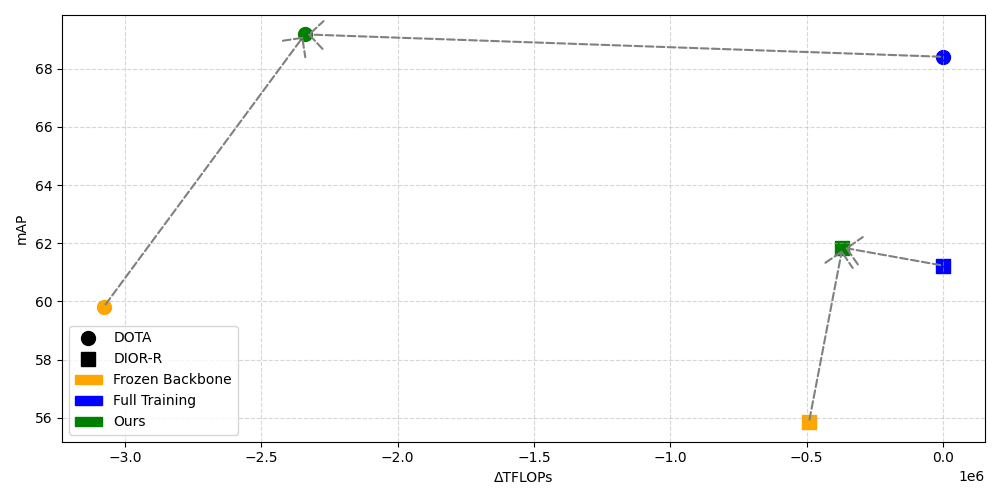}
    \caption{Impact of the proposed method on remote sensing object detection. The above visualization indicates the performance of 
    DOTA and DIOR-R benchmark datasets using models using FCOS with ResNet-50. Under long-term training, our method achieves the highest prediction accuracy (mAP) while significantly reducing computational costs (FLOPs), compared to existing methods. Besides FCOS, Faster R-CNN and RetinaNet also show similar trends.}
    \label{fig4}
\end{figure}

To tackle this limitation, we introduce a novel feature backbone fine-tuning method named \textbf{DBF} (\underline{D}ynamic \underline{B}ackbone \underline{F}reezing) for remote sensing object detection under long-term training.
     Our primary objective is to examine the performance and behavioral traits of remote sensing object detection models when subjected to prolonged training schedules with our fine-tuning approach.
    Inspired by recent studies of fine-tuning with frozen settings in natural language processing (e.g. \cite{houlsby2019parameter, li2021prefix, hu2021lora, lester2021power}), we aim to address the dilemma of whether the backbone should extract low-level generic features or specialize in domain-specific knowledge of remote sensing.
    Intuitively, we implement a module named `Freezing Scheduler' to dynamically control the update of the feature backbone for each epoch during training as shown in Fig. \ref{fig1}.
    The backbone receives knowledge concerning the downstream remote sensing data when the Freezing Scheduler sends a signal to open the backward route.
    Otherwise, the backbone becomes frozen, keeping the knowledge preserved while decreasing the computational complexity.
    Extensive experiments on DOTA \cite{xia2018dota} and DIOR-R \cite{cheng2022anchor} prove that our alternately freezing scheme can achieve better accuracy while significantly reducing GPU memory usage and training time, as indicated in Fig. \ref{fig4} and Table \ref{tab2}.
\vspace{1\baselineskip} 

\noindent\textbf{The main contributions of this work are as follows:}
\begin{itemize}
    \item We present DBF (Dynamic Backbone Freezing), a novel fine-tuning strategy that enables efficient adaptation of pre-trained models to the remote sensing domain under long-term training scenarios.
    \item DBF significantly reduces training cost—such as GPU memory usage and FLOPs—while preserving or even enhancing detection performance, especially for large-scale models for remote sensing object detection.
    \item The design of DBF is simple and framework-agnostic, ensuring seamless integration into existing remote sensing object detection models.
    \item We conduct extensive experiments on two well-known benchmarks for remote sensing object detection, including DOTA and DIOR-R, demonstrating that DBF achieves consistent gains over both full training and frozen-backbone baselines.
\end{itemize}

The remainder of this paper is organized as follows: 
    Section 2 reviews related work to provide context and highlight the contributions of this study.
    Section 3 introduces the proposed methodology, including the design and implementation of DBF. 
    Section 4 and 5 present experimental results and a detailed analysis of the method's performance. 
    Finally, Section 6 concludes the study and outlines potential directions for future research.

\section{Related work}\label{sec2}
\subsection{Pre-training and fine-tuning for remote sensing object detection}
 
Unlike natural scene images in datasets such as ImageNet, remote sensing images often exhibit distinct domain characteristics, including small object sizes, low contrast, and degradation caused by atmospheric or sensor-related conditions such as haze, fog, and rain. These factors substantially hinder detection accuracy, making image enhancement and restoration critical pre-processing steps.


Nevertheless, even with enhanced input quality, domain-specific fine-tuning remains essential, as feature backbones pre-trained on natural image datasets still require adaptation to effectively capture the unique spatial and spectral characteristics of remote sensing scenes.
 
 Modern deep learning-based object detectors usually include a feature backbone to extract key features for downstream tasks from the raw input images.
 For object detection on natural scenes, it becomes a de facto standard to initialize the backbone with pretraining on ImageNet \cite{deng2009imagenet} over the years, as it is one of the largest databases in visual object recognition, enabling models to learn a broad spectrum of features such as edges, textures, and shapes.
    Interestingly, the same research trend to use ImageNet pre-trained weights for the backbone has been observed in the remote sensing community for the following reasons.
    (i) Aerial and satellite imagery datasets often face challenges including limited diversity and scale due to geopolitical reasons, security concerns, and cost issues. For example, Wang et al. \cite{wang2022empirical} discovered the inefficiency of solely leveraging remote sensing images for detection and segmentation tasks.
    (ii) There exists an implicit consensus that though remote sensing imagery and natural scene images have different characteristics, they may still share some common visual patterns.
    (iii) ImageNet pre-trained weights are readily accessible through various deep learning frameworks online available.

Fine-tuning the backbone with datasets for downstream remote sensing tasks is a common strategy to bridge the knowledge gap between natural and remote sensing scenes, yet it may impede overall low-level feature extraction due to the domain-restricted remote sensing data, limiting performance enhancement.
    Hence, several researchers have concentrated on advanced remote sensing pre-training (e.g. \cite{fuller2022satvit, reed2023scale, zhang2023object, huang2024generic, wang2024mtp, corley2024revisiting, fuller2024croma}).

    For instance, the benefits of large unlabeled datasets for pre-training with masked autoencoding have been demonstrated \cite{fuller2022satvit}.
    Robust representations for remote sensing understanding can be learned by simultaneously acquiring general knowledge from natural images via supervised learning and domain-specific knowledge from remote sensing images through self-supervised learning \cite{huang2024generic}.
    Multi-task pre-training has also been introduced to construct remote sensing foundation models \cite{wang2024mtp}. Semi-supervised object detection using uncurated data has been proposed \cite{liu2024semi}. A transformation-invariant few-shot detection model improves generalization in limited data regimes \cite{liu2023transformation}. Point-level supervision significantly reduces annotation cost while maintaining segmentation accuracy \cite{liu2025pointsam}.

    However, we argue that the effectiveness of such pretraining strategies is still restricted in long-term training because the weights of the backbone heavily shift further from the initial pre-trained values.
    To mitigate this issue, we focus on designing a feature backbone fine-tuning method for remote sensing object detection under long-term training schemes.

\subsection{Freezing backbones in deep learning}
\label{2-a}
The frozen setting in NLP (Natural Language Processing), particularly in the context of large language models, refers to a technique where pre-trained models are used without updating their parameters for other downstream tasks.
    This research trend has gained attention for its efficiency and effectiveness in various applications.
    One significant research branch is the exploration of parameter-efficient fine-tuning methods.
    These techniques aim to fine-tune backbone models with a minimal number of additional parameters.
    For example, Adapters \cite{houlsby2019parameter}, small and trainable modules, are designed to be inserted within each layer of the pre-trained model. These are fine-tuned on specific tasks while keeping the majority of the model parameters frozen.
    On the other hand, Prefix-tuning \cite{li2021prefix} adds task-specific prefixes to the input tokens, allowing the model to adjust its behavior for different tasks without modifying the core model parameters.
    Moreover, LoRA \cite{hu2021lora} decomposes weight matrices into low-rank representations, updating only the low-rank components during fine-tuning.

In addition, recent efforts have focused on enhancing detection performance while reducing computational complexity by eliminating attention mechanisms. For example, Gao et al. \cite{
gao2023attention} propose an attention-free global multiscale fusion network tailored for remote sensing object detection, demonstrating efficient feature extraction without the computational overhead of self-attention.

Beyond parameter-efficient tuning strategies, some researchers have explored alternative architectures inspired by neuromorphic computing. For instance, Li et al. \cite{li2025brain} present a spiking neural network-based object detector, significantly reducing energy consumption while maintaining detection performance. Such approaches align with the increasing demand for lightweight and efficient models in resource-constrained remote sensing environments.

In computer vision, the usual practice for dense prediction tasks such as detection and segmentation had been to incorporate extra networks to avoid losing information in the feature backbone during fine-tuning (e.g. \cite{rebuffi2017learning, rebuffi2018efficient}).
    After the success of the frozen setting in NLP, this concept has influenced research in computer vision tasks as well.
    Lin et al. \cite{lin2022could} involves utilizing frozen pre-trained networks, with a task-specific head network to be fine-tuned for downstream tasks.
    Interestingly, Vasconcelos et al. \cite{vasconcelos2022proper} presents a fine-training scheme for detectors not to update the ImageNet-pretrained backbone. 
    This simple strategy enables models to achieve competitive performance while significantly reducing resource usage for generic object detection on natural scenes.
    However, to our knowledge, only a few have focused on the frozen setting in the same context for remote sensing understanding.
    Therefore, in this paper, we revisit the frozen setting for remote sensing object detection. 

\subsection{Recent object detection paradigm for remote sensing}
Oriented object detection refers to a specialized task where the goal is to locate and classify objects in an image while accounting for their orientation (rotation).
    Unlike traditional object detection, which uses AABBs (Axis-Aligned Bounding Boxes) to enclose objects, oriented object detection employs OBBs (Oriented Bounding Boxes) that align with the object's true orientation in the image \cite{wen2023comprehensive}.

Traditionally, convolution neural network-based oriented object detectors can be broadly classified into two categories: one-stage methods and two-stage methods.
    One-stage object detection methods frame object detection as a single regression problem. 
    These models directly predict class probabilities and bounding box coordinates from feature maps, bypassing the need for a region proposal stage. 
    By optimizing detection end-to-end, one-stage detectors achieve significantly faster inference speeds, making them highly suitable for real-time applications.
    Traditionally, these methods employ anchor-based mechanisms, where predefined anchor boxes of various scales and aspect ratios guide the network to predict bounding boxes.
    While efficient, anchor-based one-stage detectors can suffer from challenges such as anchor design sensitivity. 
To address this, anchor-free one-stage detectors, such as FCOS \cite{detector2022fcos}, Oriented Rep \cite{li2022oriented}, and CFA \cite{guo2021beyond}, propose bounding boxes by directly predicting object center points or key points, eliminating the dependency on predefined anchors. 
Two-stage object detection methods, typified by the R-CNN family (e.g., Faster R-CNN \cite{ren2016faster}, Oriented R-CNN \cite{xie2021oriented} and ReDet \cite{han2021redet}), follow a decoupled pipeline.
    The first stage generates candidate region proposals (via an RPN in anchor-based methods), while the second stage refines these proposals and predicts class labels. 
    Anchor-based two-stage detectors have long been the standard, with their region proposal networks relying on anchors to localize potential objects effectively.

Recently, transformer-based methods have emerged as a groundbreaking approach in object detection, shifting the paradigm from traditional convolutional designs to sequence modeling. These models leverage self-attention mechanisms and end-to-end optimization, as seen in DETR (DEtection TRansformer) \cite{carion2020end} and Deformable DETR \cite{zhu2020deformable}. CLT-DETR \cite{zhou2022clt} is the first adaptation of DETR for oriented object detection. AO2-DETR \cite{dai2022ao2} enhances Deformable DETR by introducing an oriented box generation and refinement module, improving localization accuracy. D2Q-DETR \cite{zhou2023d} further refines this paradigm with dynamic queries that reduce object tokens progressively, balancing accuracy and efficiency. OrientedFormer \cite{zhao2024orientedformer} introduces Gaussian positional encoding, Wasserstein self-attention, and oriented cross-attention to detect arbitrarily oriented objects in remote sensing images.

Building upon these advances, recent studies have explored efficient backbone architectures tailored to ultra-high-resolution remote sensing imagery. DynamicVis \cite{chen2025dynamicvis} proposes a selective state-space backbone that adaptively attends to informative regions while suppressing less relevant areas. This design achieves high performance with lower complexity than traditional transformer-based models, offering a promising direction for scalable and generalizable remote sensing backbones.

In addition, visual foundation models have been increasingly adopted to adapt large-scale pre-trained vision backbones to remote sensing tasks. Rather than full model retraining, these approaches often utilize efficient fine-tuning strategies. For example, RSPrompter \cite{chen2024rsprompter} builds on the Segment Anything Model (SAM), incorporating LoRA-based prompt learning for parameter-efficient instance segmentation. Similarly, RSRefSeg2 \cite{chen2025rsrefseg} introduces a CLIP-SAM integrated framework with dual-stage referring prompts for open-vocabulary segmentation. These methods demonstrate that with appropriate prompt design and cross-modal alignment, visual foundation models can be flexibly and efficiently adapted to a wide range of remote sensing applications.

\section{Proposed approach}

In this section, we introduce a simple, but efficient feature backbone fine-tuning method namely \underline{D}ynamic \underline{B}ackbone \underline{F}ine-tuning (\textbf{DBF}) for remote sensing object detection, particularly in long-term training.

\subsection{Motivation}
Compared to the existing freezing methods like \cite{vasconcelos2022proper}, which adopt a static freezing strategy, our approach introduces dynamic adjustments to the trainability of backbone parameters during training.
    This innovation enables the model to effectively balance two competing objectives in remote sensing object detection: (i) generic feature extraction, which ensures robust generalization across diverse input types (- by unfreezing -), and (ii) domain-specific feature extraction, which facilitates accurate and specialized predictions (- by freezing -).
    Previous approaches, such as \cite{vasconcelos2022proper}, often adopt a static freezing mechanism where the backbone remains either completely trainable or entirely frozen throughout the training process. 
        While this reduces computational overhead, it may fail to effectively balance generalization and specialization. 
    Fully frozen backbones risk underfitting to domain-specific characteristics, while fully trainable backbones may lose essential low-level generic features from pre-trained weights.

In this study, `long-term' training refers to schedules extending beyond 200 epochs, allowing models to gradually acquire enriched domain-specific knowledge while preserving generic features through DBF. 
    In existing papers, the models are usually trained for 12 epochs with remote sensing object detection benchmarks such as DOTA and DIOR-R \cite{li2023instance}.

\subsection{Overview of \textbf{Dynamic Backbone Freezing}}

In this work, we propose a simple but efficient feature backbone fine-tuning method for remote sensing object detection.
    Our training scheme enables the model to preserve the initialized low-level generic features from natural images while harmoniously acquiring specialized knowledge in remote sensing.
    This is achieved by alternately freezing and unfreezing the backbone using the `Freezing Scheduler' controller.

\begin{figure}[!t]
    \centering
    \includegraphics[width=14cm]{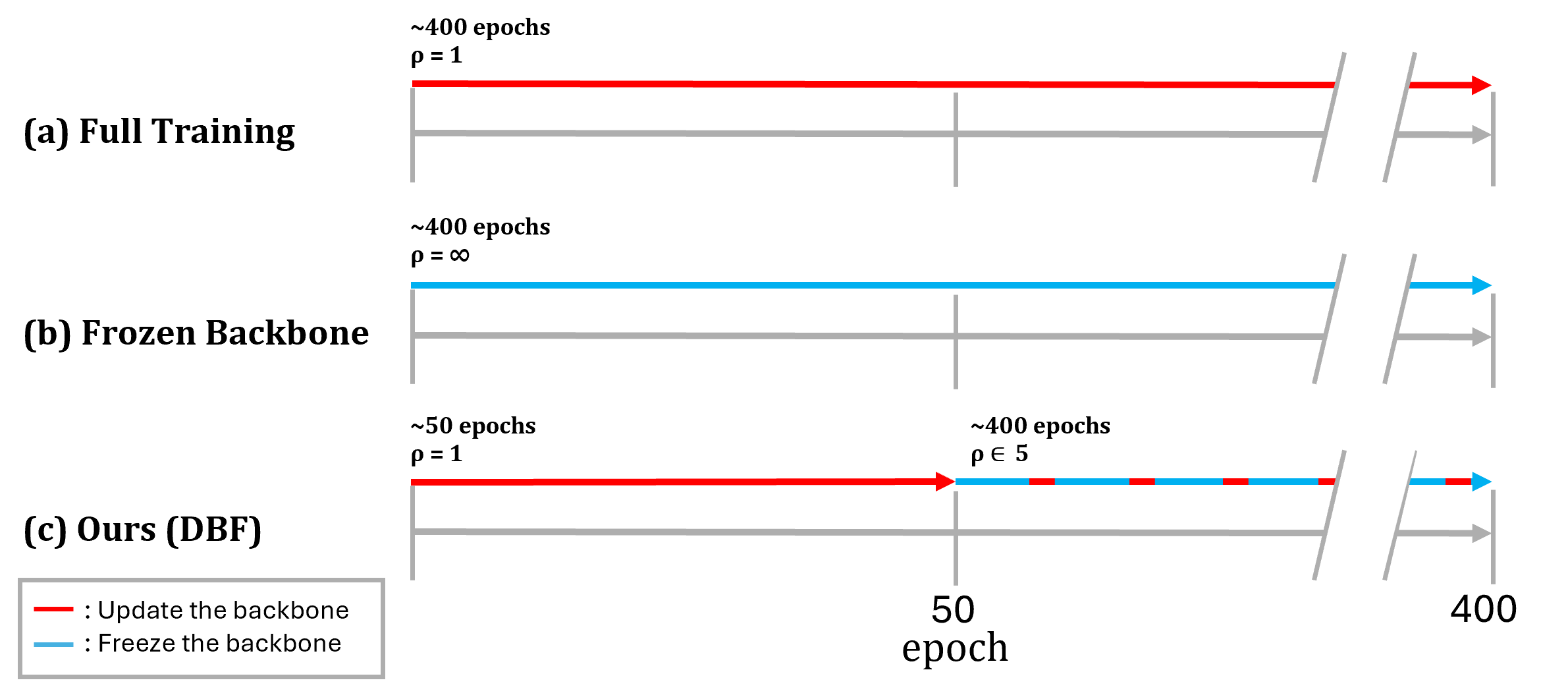}
    \caption{Visual illustration of long-term training strategies for remote sensing object detection over 400 epochs: (a) Full Training, (b) Frozen Backbone, (c) Ours (DBF).}
    \label{figex}
\end{figure}

More precisely, DBF inherently aligns with the dual objectives by:
\begin{itemize}
    \item Freezing the backbone to prioritize generalization and prevent overfitting to limited domain-specific data.
    \item Unfreezing the backbone intermittently to allow fine-tuning on specialized features as the training progresses and domain-specific patterns become more apparent.
\end{itemize}

    The detail of \textbf{DBF} is presented in Algorithm \ref{algo1}. 
    Note that in this paper all the algorithms are written in PyTorch \cite{paszke2019pytorch} style.

\lstdefinestyle{pythonstyle}{
    language=Python,
    lineskip=0.2em,
    backgroundcolor=\color{white},   
    breakatwhitespace=true, 
    commentstyle=\color{red},
    keywordstyle=\color{blue},
    basicstyle=\ttfamily\footnotesize,
    breakatwhitespace=false,         
    breaklines=true,                 
    keepspaces=true,                 
    numbers=left,       
    showspaces=false,                
    showstringspaces=false,
    showtabs=false,     
    captionpos=b,
    tabsize=2,
    numbersep=5pt,
    xleftmargin=14pt,
    frame=tb, 
    literate={ρ}{{\(\rho\)}}1, 
    escapechar=!,
}

\lstset{style=pythonstyle}

\renewcommand{\lstlistingname}{ALGORITHM}

\vspace{0.1cm}
\begin{minipage}{13cm}
\label{algo1}
\begin{lstlisting}[label=algo1,caption=Training with Dynamic Backbone Freezing.]
def forward(input):
    out = model.backbone(input)
    if freezing_scheduler(epoch): # proposed
        out = out.detach()        
    if model.neck: 
        out = model.neck(out)
    if model.roi_head:
        out = model.roi_head(out)
    out = model.bbox_head(out)
    return out
\end{lstlisting}
\end{minipage}

A deep learning-based detector is typically composed of (a) a backbone network for feature extraction, (b) a neck for feature aggregation, and (c) a detection head for predicting bounding boxes and class scores.
    In Algorithm \ref{algo1}, \texttt{backbone} and \texttt{neck} correspond to (a) and (b), respectively.
    For (c), there are two types of heads: \texttt{roi\_head} and \texttt{bbox\_head}.
    In particular, \texttt{roi\_head} is a specific type of (c) typically used in two-stage detectors like Faster R-CNN to refine the RoI (Region of Interest) proposals generated by the RPN (Region Proposal Network).
    Since some detectors may not have \texttt{neck} and \texttt{roi\_head}, conditional statements are used to handle a runtime exception in lines 5 and 7 so that the proposed method is compatible with various architectures.
    
Here, \texttt{forward()} defines the forward process of a detector model.
    It is the process of passing input data through the network to generate output.
    In this paper, we override this function for typical detector models with \texttt{freezing\_scheduler} as shown in lines 3-4.
    If the condition of line 3 is true, \texttt{out} from \texttt{backbone} becomes detached in line 4, which means gradients for the backbone will not be computed, stopping the gradient flow to the backbone during back-propagation.

    \begin{table*}[!t]
    \caption{\label{tab:table-name} Experimental configuration parameters in this work.}
    \centering
    \resizebox{\textwidth}{!}{
    \begin{tabular}{@{}l|cccc|cccc@{}}
    \toprule
    \multicolumn{1}{l|}{Dataset} & \multicolumn{4}{c|}{DOTA \cite{xia2018dota}} & \multicolumn{4}{c}{DIOR-R \cite{cheng2022anchor}} \\
    \multicolumn{1}{l|}{Backbone} & \multicolumn{2}{c}{ResNet-50 \cite{he2016deep}} & \multicolumn{2}{c|}{Swin-S \cite{liu2021swin}} & \multicolumn{2}{c}{ResNet-50 \cite{he2016deep}} & \multicolumn{2}{c}{Swin-S \cite{liu2021swin}} \\
    \multicolumn{1}{l|}{Scheduling policy} & $\sim$50 epochs & $\sim$400 epochs & $\sim$400 epochs & $\sim$800 epochs & $\sim$50 epochs & $\sim$200 epochs & $\sim$400 epochs & $\sim$1600 epochs \\ \midrule
    (a) Full training & $\rho=1$ & $\rho=1$ & $\rho=1$ & $\rho=1$ & $\rho=1$ & $\rho=1$ & $\rho=1$ & $\rho=1$ \\
    (b) Frozen Backbone \cite{vasconcelos2022proper} & $\rho=\infty$ & $\rho=\infty$ & $\rho=\infty$ & $\rho=\infty$ & $\rho=\infty$ & $\rho=\infty$ & $\rho=\infty$ & $\rho=\infty$ \\
    (c) Ours (\textbf{DBF}) & $\rho=1$ & $\rho \in \{2, 5, 10, \infty\}$ & $\rho=1$ & $\rho \in \{2, 5, 10, \infty\}$ & $\rho=1$ & $\rho \in \{2, 5, 10, \infty\}$ & $\rho=1$ & $\rho \in \{2, 5, 10, \infty\}$ \\ \bottomrule
    \end{tabular}}
    \label{tab1}
    \end{table*}

\subsection{Design choice of \textbf{Freezing Scheduler}}
\label{2-c}

The previous subsection centered on the basic pipeline of \textbf{DBF}.
    Nevertheless, we did not consider how to implement \texttt{freezing\_scheduler} in code.
    This subsection aims to discuss the design principles of `Freezing Scheduler'.

This work assumes that the input to the Freezing Scheduler is the current epoch, and the output is a binary signal indicating whether the backbone should be frozen. Specifically, a signal of 0 allows backbone updates (i.e., unfreezes the backbone), while a signal of 1 freezes it. Using this scheduler, we can dynamically control the training process by alternating between freezing and unfreezing the backbone, as illustrated in Fig. \ref{figex}. The flexibility of our framework enables researchers to design and implement various versions of the Freezing Scheduler according to their needs. Among these possible variants, we propose a simple yet effective approach called the “Step Freezing Scheduler,” which is described in Algorithm \ref{algo2}.

\vspace{0.1cm}
\begin{minipage}{13cm}
\label{algo2}
\begin{lstlisting}[label=algo2, caption=One of the simplest variants of Freezing Scheduler for \textbf{DBF} in this work: `Step Freezing Scheduler'.]
def freezing_scheduler(epoch):
    # ρ: an integer greater than 1 and
    #    less than total epochs
    if epoch % ρ == 0:
        return 0        # unfreezing
    return 1            # freezing
\end{lstlisting}
\end{minipage}

Intuitively, our Step Freezing Scheduler is similar to a step learning rate scheduler \cite{ge2019step} that decreases the learning rate at specific intervals based on the users' demand during training. 
    In Algorithm \ref{algo2}, $\rho$ is the only hyper-parameter to be tuned by users.
    Our scheduler updates the feature backbone only every $\rho$ epochs.
    In other words, a higher $\rho$ value implies that the backbone remains frozen more frequently, leading to increased efficiency in GPU memory usage and faster training time.
    Moreover, it helps to preserve the pre-trained low-level generic knowledge in the backbone while simultaneously allowing the detection model to acquire domain-specific knowledge for more precise prediction. 

\section{Experiments}
\label{sec4}
In this section, we show our novel feature backbone fine-tuning method, \textbf{DBF} enables more accurate model learning while significantly reducing computational costs under long-term training policies with CNN-based architectures for oriented object detection in remote sensing. 
    Note that CNN-based detectors can embrace the transformer as a feature backbone.

    \subsection{Experimental setup}
    \textbf{Baseline architectures.} Deep learning models for object detection are conventionally split into one-stage and two-stage methods.
        Besides, anchor-based and anchor-free methods have been recently increasingly highlighted in the classification of detector models.
        In this study, we opted for Faster R-CNN \cite{ren2016faster}, RetinaNet \cite{lin2017focal}, and FCOS \cite{detector2022fcos} as the representative models in anchor-based two-stage, anchor-based one-stage, and anchor-free one-stage object detection.
        For all benchmark models, we considered two exemplary neural networks as feature backbones: ResNet-50 \cite{he2016deep} and Swin-S \cite{liu2021swin}, one of the celebrated convolutional neural networks and transformer architectures, respectively.
        Moreover, FPN (Feature Pyramid Network)-1x \cite{lin2017feature} is used as a neck to enhance multi-scale invariance for our models.
        Here, we leverage ImageNet pre-trained weights for remote sensing object detection as they are still competitive baseline models to be readily online \cite{corley2024revisiting}.
        While alternative pre-trained models designed for remote sensing applications \cite{huang2024generic, wang2024mtp} are available, they lack the ability to promptly adapt to novel network architectures.
        That is, for newly proposed neural networks, the weights trained on natural image datasets are more easily accessible. In contrast, those specialized in remote sensing imagery take considerable time to deploy.
        Thus, our experiments explore when the ImageNet pre-trained weights meet our method to boost remote sensing object detection.

\textbf{Training details.} We chose PyTorch and MMDetection \cite{chen2019mmdetection} / MMRotate \cite{zhou2022mmrotate} for implementing our method and baseline architectures.
        For model optimization, SGD (Stochastic Gradient Descent) \cite{amari1993backpropagation, zinkevich2010parallelized} is used with momentum of weight 0.9, weight decay 0.0001, and batch size 8.
        The initial learning rate is set to 0.005 for all detectors.
        To enhance the stability and convergence of our training process, we utilize a linear learning rate warm-up strategy \cite{kalra2024warmup} over the initial 500 iterations, during which the learning rate increases linearly from zero to 33\% of the maximum learning rate. 
        The learning rate decayed by 1/4 after the 12th epoch.
        To prevent gradient explosion, L2 norm-based gradient clipping \cite{zhang2019gradient} is used with a maximum norm of 35.
        We apply the following data transformations for training: `\footnote{RResize is a rotation-aware resizing operation designed for resizing images and their associated oriented bounding boxes. For DOTA, it adapts the input to a predefined size, 1024x1024 while maintaining the aspect ratio. Similarly, for DIOR-R, it adjusts all inputs to a uniform target size, 800x800 while maintaining the aspect ratio. RResize is critical for efficient batching and optimized memory usage.}RResize', `\footnote{RRandomFlip is a rotation-aware data augmentation technique that randomly flips images along specified axes (horizontal, vertical, or diagonal directions) while ensuring that the corresponding oriented bounding boxes are updated accordingly. For both DOTA and DIOR datasets, we consider all three directions with probability 0.25 in RRandomFlip.}RRandomFlip', and `\footnote{All images are normalized per each color channel (R/G/B) with means [123.68, 116.28, 103.53] and standard deviations [58.40, 57.12, 57.38].}Normalize' \cite{zhou2022mmrotate}. 
        For a fair comparison, we set the value of the random seed to 0 during experiments.

  To demonstrate the effectiveness of the proposed method, we compare it against (a) full training and (b) `Frozen Backbone' \cite{vasconcelos2022proper}. 
        In (a), the backbone is always fine-tuned every training epoch, which requires the highest computing resources.
        Meanwhile, in (b), the backbone is never fine-tuned (i.e. always frozen), leading to minimal resource usage.
        Our method balances the extremes of (a) and (b).
        Precisely, our approach can be seen as a generalized version of (a) and (b): 
        \begin{itemize}
            \small \item If $\rho$ equals 1, our method is equivalent to (a); \vspace{-0cm}
            \item If $\rho$ equals $\infty$, our method is equivalent to (b).
        \end{itemize}

    Table \ref{tab1} indicates the experimental configuration parameters in this work.
    As mentioned in Sec. \ref{2-c}, we use Step Freezing Scheduler for \textbf{DBF}.
    Besides, we consider four $\rho$ values, $\{2, 5, 10, \infty\}$ in our method: the models fully fine-tuned (i.e. $\rho=1$) for the first 50 or 400 epochs are fine-tuned again with \textbf{DBF} under the four $\rho$ values.

\begin{table*}[!t]
    \caption{\label{tab:table-name} Experimental results on DOTA and DIOR-R with various convolutional object detection architectures (\textbf{Bold} text indicates the best while \underline{underlined} text means the second-best in this paper). Notably, our method achieves higher validation mAP scores across nearly all cases for both datasets compared to approaches (a) and (b), underscoring its efficacy in long-term training scenarios for CNN-based object detection models in remote sensing.}
\centering
\resizebox{14cm}{!}{%
\begin{tabular}{@{}cccccccccccccc@{}}
\toprule
 &  &  & \multicolumn{2}{c}{} & \multicolumn{1}{c}{} & \multicolumn{8}{c}{\cellcolor[HTML]{FFCE93}(c) Ours (\textbf{DBF})} \\
 &  &  & \multicolumn{1}{c}{\multirow{-2}{*}{ \makecell{(a) Full training\\(Baseline)} }} & \multicolumn{2}{c}{\multirow{-2}{*}{ \makecell{ \makecell{(b) Frozen Backbone\\\cite{vasconcelos2022proper}}  } }} & \multicolumn{2}{c}{\cellcolor[HTML]{FFCE93}\textcolor{gray}{($\rho=1 \rightarrow$)} $\rho=2$} & \multicolumn{2}{c}{\cellcolor[HTML]{FFCE93}\textcolor{gray}{($\rho=1 \rightarrow$)} $\rho=5$} & \multicolumn{2}{c}{\cellcolor[HTML]{FFCE93}\textcolor{gray}{($\rho=1 \rightarrow$)} $\rho=10$} & \multicolumn{2}{c}{\cellcolor[HTML]{FFCE93}\textcolor{gray}{($\rho=1 \rightarrow$)} $\rho=\infty$} \\ \cmidrule(l){7-14} 
\multirow{-3.5}{*}{Dataset}& \multirow{-3.5}{*}{Detector}& \multirow{-3.5} {*}{Backbone}& mAP & mAP & $\Delta$TFLOPs & mAP & $\Delta$TFLOPs & mAP & $\Delta$TFLOPs & mAP & $\Delta$TFLOPs & mAP & $\Delta$TFLOPs \\ \midrule
 &  & \makecell{ResNet-50\\\cite{he2016deep}} & 69.00 & \makecell{61.31\\(\textcolor{blue}{-7.69})} & \textbf{-3,077,501} & \makecell{69.03\\(\textcolor{red}{+0.03})} & -1,170,338 & \makecell{69.22\\(\textcolor{red}{+0.22})} & -1,872,541 & \makecell{\underline{69.27}\\(\textcolor{red}{+0.27})} & -2,106,608 & \makecell{\textbf{69.28}\\(\textcolor{red}{+0.28})} & \underline{-2,340,676} \\ \cmidrule(l){3-14} 
 & \multirow{-3.5}{*}{\makecell{Faster R-CNN\\\cite{ren2016faster}}} & \makecell{Swin-S\\\cite{liu2021swin}} & 71.23  & \makecell{68.14\\(\textcolor{blue}{-3.09})}  & \textbf{-6,115,002} & \makecell{\textbf{71.80}\\(\textcolor{red}{+0.57})} & -1,538,750 & \makecell{71.73\\(\textcolor{red}{+0.50})} & -2,462,001 & \makecell{\underline{71.74}\\(\textcolor{red}{+0.51})} & -2,769,751 &  \makecell{71.73\\(\textcolor{red}{+0.50})} & \underline{-3,077,501} \\ \cmidrule(l){2-14} 

 &  & \makecell{ResNet-50\\\cite{he2016deep}} & 64.19 & \makecell{57.29\\(\textcolor{blue}{-6.90})}  & \textbf{-3,077,501} & \makecell{64.48\\(\textcolor{red}{+0.29})}  & -1,170,338  & \makecell{\underline{64.89}\\(\textcolor{red}{+0.70})}  & -1,872,541  & \makecell{64.77\\(\textcolor{red}{+0.58})} & -2,106,608 & \makecell{\textbf{64.98}\\(\textcolor{red}{+0.79})} & \underline{-2,340,676}  \\ \cmidrule(l){3-14} 
 & \multirow{-3.5}{*}{\makecell{RetinaNet\\\cite{lin2017focal}}} & \makecell{Swin-S\\\cite{liu2021swin}} & 64.89 & \makecell{61.61\\(\textcolor{blue}{-3.28})} & \textbf{-6,115,002} & \makecell{64.92\\(\textcolor{red}{+0.03})} & -1,538,750 & \makecell{\textbf{65.02}\\(\textcolor{red}{+0.13})} & -2,462,001 & \makecell{\underline{65.01}\\(\textcolor{red}{+0.12})} & -2,769,751 & \makecell{64.97\\(\textcolor{red}{+0.08})} & \underline{-3,077,501} \\ \cmidrule(l){2-14} 

 &  & \makecell{ResNet-50\\\cite{he2016deep}} & 68.41 & \makecell{59.82\\(\textcolor{blue}{-8.59})} & \textbf{-3,077,501} & \makecell{68.41\\(\textcolor{black}{+0.00})}  & -1,170,338 & \makecell{68.80\\(\textcolor{red}{+0.39})} & -1,872,541 & \makecell{\underline{68.98}\\(\textcolor{red}{+0.57})} & -2,106,608 & \makecell{\textbf{69.18}\\(\textcolor{red}{+0.77})} & \underline{-2,340,676}  \\ \cmidrule(l){3-14} 
 \multirow{-13.7}{*}{\makecell{DOTA\\ \cite{xia2018dota}}} & \multirow{-3.5}{*}{\makecell{FCOS\\\cite{detector2022fcos}}} & \makecell{Swin-S\\\cite{liu2021swin}} & 71.47 & \makecell{67.45\\(\textcolor{blue}{-4.02})} &  \textbf{-6,115,002} & \makecell{71.46\\(\textcolor{blue}{-0.01})} & -1,538,750 & \makecell{71.44\\(\textcolor{blue}{-0.03})} & -2,462,001 & \makecell{\textbf{71.62}\\(\textcolor{red}{+0.15})} & -2,769,751 & \makecell{\underline{71.51}\\(\textcolor{red}{+0.04})} & \underline{-3,077,501} \\ \midrule

 &  & \makecell{ResNet-50\\\cite{he2016deep}} & 63.82 & \makecell{54.17\\(\textcolor{blue}{-9.65})}  & \textbf{-491,825}  & \makecell{\textbf{64.13}\\(\textcolor{red}{+0.31})} & -184,434 & \makecell{63.85\\(\textcolor{red}{+0.03})} & -295,095 & \makecell{63.86\\(\textcolor{red}{+0.04})}  & -331,982 & \makecell{\underline{63.96}\\(\textcolor{red}{+0.14})} & \underline{-368,869} \\ \cmidrule(l){3-14} 
 & \multirow{-3.5}{*}{\makecell{Faster R-CNN\\\cite{ren2016faster}}} & \makecell{Swin-S\\\cite{liu2021swin}} & 67.68 & \makecell{62.64\\(\textcolor{blue}{-5.04})}  & \textbf{-4,588,368} &  \makecell{67.97\\(\textcolor{red}{+0.29})} & -1,720,638  &  \makecell{67.99\\(\textcolor{red}{+0.31})} & -2,753,021 & \makecell{\textbf{68.06}\\(\textcolor{red}{+0.38})}  & -3,097,149 & \makecell{\underline{68.03}\\(\textcolor{red}{+0.35})} & \underline{-3,441,276} \\ \cmidrule(l){2-14} 
 
 &  & \makecell{ResNet-50\\\cite{he2016deep}} & 59.09  & \makecell{53.88\\(\textcolor{blue}{-5.21})} & \textbf{-491,825} & \makecell{\textbf{59.26}\\(\textcolor{red}{+0.17})} & -184,434  & \makecell{\underline{59.25}\\(\textcolor{red}{+0.16})}  &  -295,095 & \makecell{58.94\\(\textcolor{blue}{-0.15})} & -331,982 & \makecell{58.97\\(\textcolor{blue}{-0.12})} & \underline{-368,869}  \\ \cmidrule(l){3-14} 
 & \multirow{-3.5}{*}{\makecell{RetinaNet\\\cite{lin2017focal}}} & \makecell{Swin-S\\\cite{liu2021swin}} & 65.11  & \makecell{60.81\\(\textcolor{blue}{-4.30})}   & \textbf{-4,588,368} & \makecell{\textbf{65.15}\\(\textcolor{red}{+0.04})} & -1,720,638 & \makecell{\underline{65.14}\\(\textcolor{red}{+0.03})} & -2,753,021 & \makecell{64.86\\(\textcolor{blue}{-0.25})} & -3,097,149 & \makecell{64.75\\(\textcolor{blue}{-0.36})} & \underline{-3,441,276} \\ \cmidrule(l){2-14}

 &  & \makecell{ResNet-50\\\cite{he2016deep}} & 61.23 & \makecell{55.84\\(\textcolor{blue}{-5.39})}  & \textbf{-491,825}  & \makecell{61.39\\(\textcolor{red}{+0.16})} & -184,434 & \makecell{61.63\\(\textcolor{red}{+0.40})}  & -295,095 & \makecell{\underline{61.66}\\(\textcolor{red}{+0.43})} & -331,982  & \makecell{\textbf{61.85}\\(\textcolor{red}{+0.62})} & \underline{-368,869} \\ \cmidrule(l){3-14} 
\multirow{-13.7}{*}{\makecell{DIOR-R\\ \cite{cheng2022anchor} }} & \multirow{-3.5}{*}{\makecell{FCOS\\\cite{detector2022fcos}}} & \makecell{Swin-S\\\cite{liu2021swin}} & \makecell{67.58} & \makecell{67.48\\(\textcolor{blue}{-0.10})} & \textbf{-4,588,368} & \makecell{67.69\\(\textcolor{red}{+0.11})} & -1,720,638 & \makecell{\textbf{68.08}\\(\textcolor{red}{+0.50})} & -2,753,021 & \makecell{\underline{68.05}\\(\textcolor{red}{+0.47})} & -3,097,149 & \makecell{\underline{68.05}\\(\textcolor{red}{+0.47})} & \underline{-3,441,276} \\ \bottomrule
\end{tabular}%
}
\label{tab2}
\end{table*}

\begin{figure*}[t]
    \centering
    \includegraphics[width=13cm]{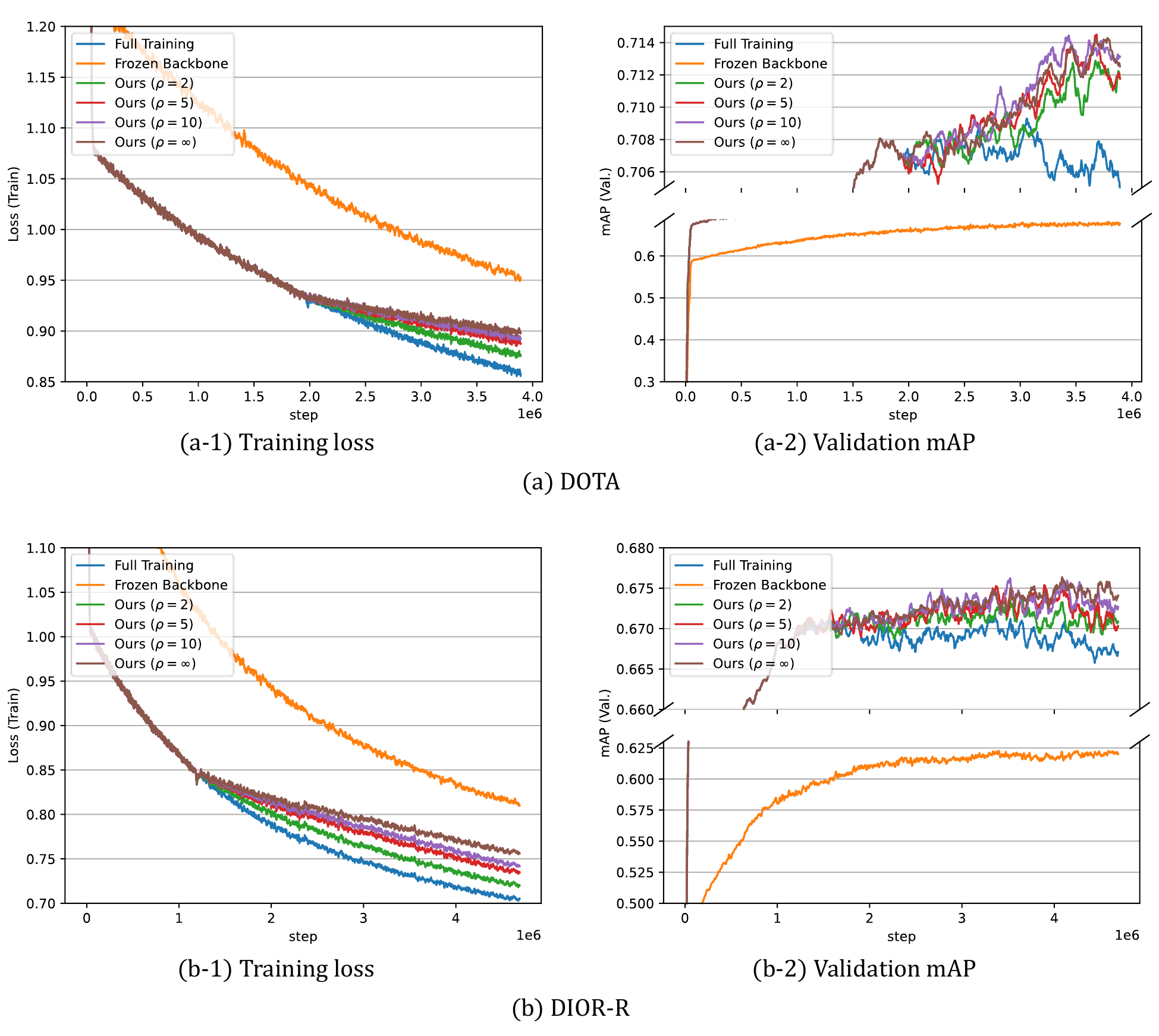}
    \vspace{-0.2cm}
    \caption{Learning curves on DOTA and DIOR-R with Faster R-CNN and Swin-S: It can be seen that while our \textbf{DBF} shows slower training loss reduction than `Full training', it shows stable and highest validation mAP values. 
    Particularly, the validation mAP consistently shows an upward trend when using our method under long-term training scenarios.
    In the case of `Frozen Backbone', while it incurs the lowest training cost, there is a considerable degradation in model performance compared to other approaches.}
    \label{fig2}
\end{figure*}

   \subsection{Evaluation metrics}
   \label{eval_metrics}
   
    \textbf{mAP (mean Average Precision)} is a key metric used to evaluate the accuracy of object localization and classification.
    We adopt mAP to provide a balanced assessment of models with both precision and recall across different categories in multiple object detection.
        Assume that TP means true positives, FP means false positives, and FN means false negatives. 
        Precision is the proportion of predicted objects that are correctly identified as follows:
         \begin{equation}
             {\text{Precision}} = \frac{\text{TP}}{\text{TP+FP}}.
        \end{equation}

        Recall is the proportion of actual objects correctly detected by the model as:
        \begin{equation}
             {\text{Recall}} = \frac{\text{TP}}{\text{TP+FN}}.
        \end{equation}
    We recommend to see \cite{wang2022parallel} for detailed calculation of mAP.
    
    Precisely, `mAP@50' \cite{everingham2015pascal} is used for model evaluation following \cite{li2023instance}.
        In mAP@50, predictions with IoU $\geq$ 0.5 are considered correct, where IoU (Intersection over Union) per each prediction is calculated as follows:
        \begin{equation}
             {\text{IoU}} = \frac{\text{Area of overlap between $\hat{y}$ and $y$}}{\text{Area of union}},
        \end{equation}
        where $\hat{y}$ is the predicted bounding box and $y$ is the ground truth bounding box.

    \textbf{FLOPs (FLoating point OPerations)} refers to the total number of floating-point operations performed by a model during training or inference. 
        It represents the computational cost of a model.    
        Matrix multiplications and additions contribute to the FLOPs count during input propagation through the model.
    Typically, FLOPs are calculated based solely on the one forward pass with a batch size of 1 \cite{johnson2018rethinking}.

    On the other side, we use ${\text{F}_{\text{total}}}$ instead as follows:
    \vspace{-0.1cm}
    \begin{equation}\label{eq1}
         {\text{F}_{\text{total}}} = \sum_{\text{e}=1}^{\text{E}} \sum_{\text{i}=1}^{\text{N}} \sum_{\text{l}=1}^{\text{L}} \text{(} {\text{F}_{\text{forward}}}\text{(e, l)} + {\text{F}_{\text{backward}}}\text{(e, l)} \text{)},
        \vspace{-0.1cm}
        \end{equation}
        where $\text{E}$ is total training epochs, $\text{N}$ is the number of training samples, and $\text{L}$ is the number of network layers. ${\text{F}_{\text{forward}}}\text{(e, l)}$ and ${\text{F}_{\text{backward}}}\text{(e, l)}$ correspond to FLOPs for forward and backward operations at the $\text{e}$-th epoch in the $\text{l}$-th layer, respectively.
        During computation, ${\text{F}_{\text{forward}}}$ only focuses on the input, whereas ${\text{F}_{\text{backward}}}$ takes into account the weights and inputs for gradient flow.
        Hence, this work assumes that ${\text{F}_{\text{backward}}}$ is twice that of ${\text{F}_{\text{forward}}}$.
        Intuitively, $\Delta{\text{F}_{\text{total}}}$ effectively shows how much computation reduces during all training epochs.
    
    \subsection{Datasets}
        We perform experiments on the following datasets for generic multiple object detection in remote sensing: DOTA-v1.0 \cite{xia2018dota} and DIOR-R \cite{cheng2022anchor}. 
        For our experiments, we only consider labels for oriented object detection, as detecting rotated objects is more practical in remote sensing.
        DOTA consists of 15 object categories including `car', `ship', and `harbor', with 2,806 aerial images and 188,282 bounding box labels.
        DIOR-R surpasses DOTA with its 20 object categories and 192,512 instances captured in 23,463 images.
        Note that we adopt standard training and test protocols for both datasets following prior researches like \cite{li2023instance}.

  \subsection{Experimental results}
    As seen in Table \ref{tab2}, our method performs better in most cases for all datasets.
    Here, $\Delta$TFLOPs denotes the difference in Tera FLOPs relative to the baseline (a).
    In other words, the lower $\Delta$TFLOPs, the better.
    As mentioned before, the FLOPs should be calculated as ${\text{F}}_{\text{total}}$ in Eq. \ref{eq1}.
    Besides, `($\rho=1 \rightarrow$) $\rho=\alpha$' in (c) means that the backbone is fine-tuned with $\rho=1$ for the first 50 or 400 epochs and then fine-tuned again with $\rho=\alpha$ according to Table \ref{tab1}.
    Although (a) permits adjusting the backbone for every epoch, it mostly results in slightly lower mAP than ours.
    Besides, (b) decreases $\Delta$TFLOPs the most, but it undergoes a significant drop in mAP (about 3 to 10\%p).
    Compared to (a) and (b), the proposed approach balances low time complexity and precise inference ability: our method achieves the highest mAP for almost all cases, while effectively reducing computational costs during training.

To discuss the practical effectiveness of the proposed technique, we present how much training time can be saved when adopting our approach.
    Let the detector be FCOS with ResNet-50 as a backbone.
    For the training split of DOTA, using the Nvidia RTX A5000, it takes an average of 23 minutes to fully fine-tune the model, including the backbone, in one epoch. 
    Freezing the backbone takes 16 minutes on average.
    Based on these, we can approximately calculate the training time for each training policy as shown in Table \ref{tab3}.

Here, $\Delta$T means the difference in training time compared to (a).
    Interestingly, ours with $\rho=\infty$ achieves a 0.77 higher mAP while reducing the training time by 2,450 minutes (1.7 days) compared to the baseline.
    Reducing model learning time by several days or months holds significant benefits as:
\begin{itemize}
     \item \textit{Cost reduction}: Cutting down on training time can significantly lower cloud usage costs and power bills; 
    \item \textit{Productivity improvement}: Shorter training times enable more experiments (e.g. hyper-parameter tuning or ablation study) to be conducted quickly; 
    \item \textit{Sustainable AI}: Lower power consumption for model training translates to reduced carbon emissions, leading to environmental conservation and eco-friendly AI. 
\end{itemize}

\begin{table}[]
        \caption{\label{tab:table-name} Examples of training time savings on Nvidia RTX A5000. Although long-term training usually entails significant computational time and memory usage, the proposed approach enables a more efficient process with reduced resource requirements. Remarkably, it achieves this while delivering superior inference accuracy.}
\centering
\begin{tabular}{@{}ccccccc@{}}
\toprule
 & \multicolumn{2}{c}{Training policy} &  &  &  &  \\ \cmidrule(lr){2-3}
\multirow{-2}{*}{\begin{tabular}[c]{@{}c@{}}FCOS + \\ ResNet-50 on DOTA\end{tabular}} & \begin{tabular}[c]{@{}c@{}}$\sim$50\\ epochs\end{tabular} & \begin{tabular}[c]{@{}c@{}}$\sim$400\\ epochs\end{tabular} & \multirow{-2}{*}{mAP} & \multirow{-2}{*}{$\Delta$ TFLOPs} & \multirow{-2}{*}{\begin{tabular}[c]{@{}c@{}}Training Time\\(\textit{T})\end{tabular}} & \multirow{-2}{*}{$\Delta$ \textit{T}} \\ \midrule
(a) Full training & $\rho=1$ & $\rho=1$ & 68.41 & - & 9,200m & - \\
(b) Frozen Backbone & $\rho=\infty$ & $\rho=\infty$ & 59.82 & \textbf{-3,077,501} & \textbf{6,400m} & \textbf{-2,800m} \\
\rowcolor[HTML]{FFCE93} 
\cellcolor[HTML]{FFCE93} & $\rho=1$ & $\rho=2$ & 68.41 & -1,170,338 & 7,975m & -1,225m \\
\rowcolor[HTML]{FFCE93} 
\cellcolor[HTML]{FFCE93} & $\rho=1$ & $\rho=5$ & 68.80 & -1,872,541 & 7,240m & -1,960m \\
\rowcolor[HTML]{FFCE93} 
\cellcolor[HTML]{FFCE93} & $\rho=1$ & $\rho=10$ & \underline{68.98} & -2,106,608 & 6,995m & -2,205m \\
\rowcolor[HTML]{FFCE93} 
\multirow{-4}{*}{\cellcolor[HTML]{FFCE93}(c) Ours} & $\rho=1$ & $\rho=\infty$ & \textbf{69.18} & {\underline{-2,340,676}} & {\underline{6,750m}} & {\underline {-2,450m}} \\ \bottomrule
\end{tabular} \label{tab3}  %
\end{table}

   Fig. \ref{fig2} further demonstrates the effectiveness of the proposed method.
    In all subfigures, a downward trend is visible in the validation mAP curves for (a) full training, while the training loss also decreases the fastest, indicating over-fitting.
    On the other hand, the proposed method gradually reduces the training loss while improving the validation mAP, or at least preventing it from decreasing, especially for higher $\rho$.
    These observations show that the proposed method helps in stable model convergence while reducing memory usage.

    \subsection{Parameter study of $\rho$}
    This subsection examines how users can effectively select $\rho$ value.
    In the experiments, we consider four $\rho$ values, $\{2, 5, 10, \infty\}$ for our method.
    One certain thing is that as the $\rho$ value increases, more time and memory are saved in training.
    However, larger $\rho$ deprives the learning opportunity of the feature backbone, which might hurt the model generalization.
    To find the best $\rho$, we compute the average of $\Delta$mAP over all the experiments for each $\rho$.
    Table \ref{tab4} indicates that $\rho=\infty$ is the optimal choice and the correlation between $\rho$ and the improvement in model performance is evident.

    \begin{table}[h]
        \caption{\label{tab:table-name} Average of $\Delta$mAP for each $\rho$ in \textbf{DBF}.}
    \centering
    \begin{tabular}{@{}ccccc@{}}
    \toprule
     $\Delta$mAP & $\rho=2$ & $\rho=5$ & $\rho=10$ & $\rho=\infty$ \\ \midrule
    mean ± standard deviation & 0.17±0.2 & \underline{0.28}±0.2 & 0.27±0.3 & \textbf{0.30}±0.3 \\ \bottomrule
    \end{tabular}%
    \label{tab4}
    \end{table}

     \subsection{Weakness analysis}

\begin{figure*}[t]
    \centering
    \includegraphics[width=13cm]{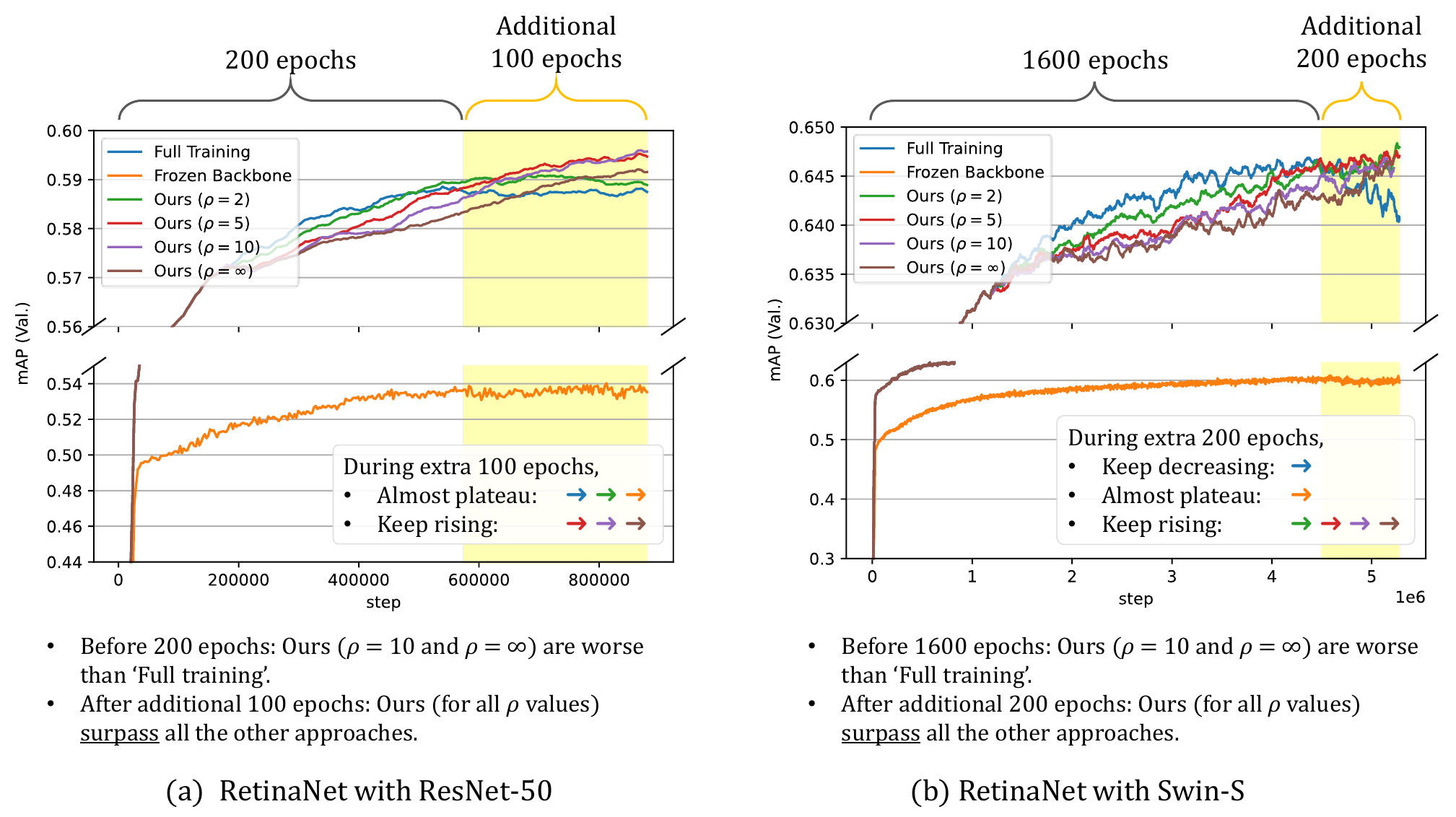}
    \caption{Effects of further training of the proposed method for RetinaNet on DIOR-R: It can be confirmed that though our \textbf{DBF} achieves worse mAP than the baseline when $\rho \in \{10, \infty\}$ during the initial training following the policies in Table \ref{tab1}, the prediction accuracy of ours are eventually improved with longer additional training. Besides, it brings steady performance improvement even in long-term training, compared to other methods.}
    \label{fig3}
\end{figure*}
    
    Although our approach usually achieves competitive performance on several benchmarks and detectors, a relatively larger performance drop (-0.12 to -0.36) can be observed on the DIOR-R dataset when $\rho$ is greater than 10 for the RetinaNet architecture as shown in Table \ref{tab1}.
        Fortunately, this is not a limitation of the proposed method but because we terminate training early following the pre-fixed policy in Table \ref{tab1}, even though the model has not converged sufficiently.
        As shown in Fig \ref{fig3}, our method surpasses the other approaches for all $\rho$ values when we additionally train the models with ResNet-50 and Swin-S for 100 and 200 epochs more, respectively.
        Remark that we attempted to train more than 100 or 200 epochs for each case; but, we observed that the performance of all methods either deteriorated or showed no improvement, so we ceased further training.

    Despite the success of our method, current work has notable weaknesses. First, the optimal $\rho$ value is not always fixed.
        As indicated in Table \ref{tab2}, $\rho=\infty$ is not the optimal for some cases.
        For instance, in Faster R-CNN with ResNet-50, the mean Average Precision gradually increases as $\rho$ goes to $\infty$ for DOTA (+0.03 $\rightarrow$ +0.22 $\rightarrow$ +0.27 $\rightarrow$ 0.28), whereas for DIOR-R, a U-shaped mAP trend is observed (+0.31 $\rightarrow$ +0.03 $\rightarrow$ +0.04 $\rightarrow$ +0.14).
        To address this, future work would include exploring more diverse freezing schedulers that automatically decide whether to update the backbone based on input samples or learning curves online.
        
    Besides, there is still room for further improvement in our method, in terms of training cost.
        For example, Adapters \cite{houlsby2019parameter} for NLP can be leveraged in our method to boost more efficient fine-tuning without updating the entire backbone.

\section{Comparison with state-of-the-art methods}
In Section \ref{sec4}, DBF has proven to be an efficient fine-tuning strategy for long-term training in CNN-based remote sensing object detection models.
    To further validate its effectiveness, we extend our experiments by incorporating DBF into a state-of-the-art end-to-end transformer-based object detection model, OrientedFormer \cite{zhao2024orientedformer}.
    In this section, we also include a comparison with state-of-the-art methods, for comparative evaluation.

\begin{table}[!h]
    \centering
    \caption{\label{tab:table-name} Comparison with state-of-the-art oriented object detection models on DIOR-R. Here, OrientedFormer, combined with our training scheme, achieves the best mAP, demonstrating the effectiveness of the proposed method.}
    \vspace{0.2cm} 
    \begin{tabular}{@{}lcl@{}}
        \toprule
        \textbf{Detector} & \textbf{Backbone} & \textbf{mAP} \\
        \midrule
        \multicolumn{3}{l}{\textbf{One-stage:}} \\
        RetinaNet (ICCV 2017) \cite{lin2017focal} & ResNet-50 \cite{he2016deep} & 57.55 \\
        DFNet (IEEE TGRS 2024) \cite{xie2024oriented} & ResNet-50 \cite{he2016deep} & 62.11 \\
        Oriented Rep (CVPR 2022) \cite{li2022oriented} & ResNet-50 \cite{he2016deep} & 66.71 \\
        DCFL (CVPR 2023) \cite{xu2023dynamic} & ResNet-50 \cite{he2016deep} & 66.80 \\
        \midrule
        \multicolumn{3}{l}{\textbf{Two-stage:}} \\
        Gliding Vertex (IEEE TPAMI 2020) \cite{xu2020gliding} & ResNet-50 \cite{he2016deep} & 60.06 \\
        RoI Transformer (CVPR 2019) \cite{ding2019learning} & ResNet-50 \cite{he2016deep} & 63.87 \\
        QPDet (IEEE TGRS 2023) \cite{yao2022improving} & ResNet-50 \cite{he2016deep} & 64.20 \\
        AOPG (IEEE TGRS 2022) \cite{cheng2022anchor} & ResNet-50 \cite{he2016deep} & 64.41 \\
        DODet (IEEE TGRS 2022) \cite{cheng2022dual} & ResNet-50 \cite{he2016deep} & 65.10 \\
        \midrule
        \multicolumn{3}{l}{\textbf{End-to-end:}} \\
        ARS-DETR (IEEE TGRS 2024) \cite{zeng2024ars} & ResNet-50 \cite{he2016deep} & 66.12 \\
        OrientedFormer (IEEE TGRS 2024) \cite{zhao2024orientedformer} & Swin-S \cite{liu2021swin} & \underline{68.84} \\
        \rowcolor[HTML]{FFCE93}
        \textbf{OrientedFormer \cite{zhao2024orientedformer} + DBF (Ours)} & Swin-S \cite{liu2021swin} & \textbf{69.38} (\textcolor{red}{+0.54}) \\
        \bottomrule
    \end{tabular}
    \vspace{0.2cm}
    \label{tab5}
\end{table}

\subsection{Experimental setup}
In this section, we choose OrientedFormer \cite{zhao2024orientedformer} as the representative end-to-end transformer-based object detection models in remote sensing.
    In addition, we opt for a DIOR-R dataset, one of the well-known large-scale datasets for oriented object detection in remote sensing.
        Following \cite{zhao2024orientedformer}, we use the same data augmentation strategies for DIOR-R, as mentioned in Section \ref{sec4}.
    To show the accuracy of our method, we use `mAP@50' as an evaluation metric, as in Section \ref{eval_metrics}.
    In the original paper \cite{zhao2024orientedformer}, the model is trained for 12 epochs with the image size 800x800 with DIOR-R.
        However, we train the model for 250 epochs with the proposed fine-tuning approach. 
        In the previous section, we observed `($\rho=1 \rightarrow$) $\rho=\infty$' in DBF is the optimal choice under long-term training.
        Hence, we set $\rho$ as 1 for first 50 epochs, and then $\infty$ for the remaining 200 epochs.

\subsection{Experimental results}
We compare our approach against both recent CNN-based and transformer-based detectors on DIOR-R.
    As indicated in Table \ref{tab5}, our approach achieves 69.38\%, surpassing all compared CNN-based one-stage and two-stage detectors as well as end-to-end transformer-based detectors.
        Specifically, our method improves upon the baseline OrientedFormer model by +0.54 mAP, demonstrating the effectiveness of integrating Dynamic Backbone Freezing (DBF) for enhanced performance in remote sensing object detection tasks.

\section{Conclusions and future work}

In this paper, we have proposed a simple but efficient backbone fine-tuning strategy named DBF for the remote sensing object detection.     
    This work demonstrated that alternating between freezing and unfreezing the backbone aids remote sensing object detection models in learning both general and domain-specific knowledge under long-term training. 
        Particularly with the proposed method, a significant decrease in memory usage and training time can be achieved under very long training schedules. 
    Our approach has shown powerful results on DOTA and DIOR-R datasets with various detection architectures. 
Additionally, the intuitive design of DBF guarantees high compatibility with current deep learning frameworks, facilitating easy comprehension and seamless integration.

    In the future, more sophisticated studies will be conducted as follows: 
   (i) connecting the current method with the backbone initialized with other remote sensing pre-training methodologies like \cite{huang2024generic, wang2024mtp} rather than ImageNet-based one; 
   (ii) extending this work to other remote sensing understanding tasks such as segmentation and change detection;
   (iii) designing novel freezing schedulers to mitigate class imbalance issues \cite{kim2021imbalanced} under long-term training.
    We believe these research directions are crucial for future work.


\bmhead{Author contributions}
JP and YK: Data curation, conceptualization, methodology and experimenting. MJ: Methodology and supervision.

\bmhead{Funding}

This work was supported by the Agency For Defense Development Grant funded by the Korean Government (UI220066WD).

\bmhead{Availability of data and materials}

Not applicable.

\bmhead{Code availability}

Our code will be released on GitHub after our paper is accepted.

\section*{Declarations}

\bmhead{Ethics approval and consent to participate}
Not applicable.

\bmhead{Consent for publication}
Not applicable.

\bmhead{Competing interests}

The authors declare that they have no known competing financial interests or personal relationships that could have appeared to influence the work reported in this paper.



\end{document}